\documentclass[10pt, conference, compsocconf]{IEEEtran}
\ifCLASSINFOpdf
\else
\fi

\usepackage{graphicx}
\usepackage{cite}
\usepackage{amsmath,amssymb,amsfonts}
\usepackage{hyperref}

\begin{document}
\title{Unsupervised Image-to-Image Translation with Self-Attention Networks}

% author names and affiliations
% use a multiple column layout for up to two different
% affiliations

\author{\IEEEauthorblockN{Taewon Kang}
\IEEEauthorblockA{Sejong Academy of Science and Arts\\
Sejong, Korea\\
itschool@itsc.kr}
\and
\IEEEauthorblockN{Kwang Hee Lee*} 
\IEEEauthorblockA{Boeing Korea Engineering and \\ Technology Center (BKETC)\\
Seoul, Korea\\
kwanghee.lee2@boeing.com\\
* indicates corresponding author}
}

% conference papers do not typically use \thanks and this command
% is locked out in conference mode. If really needed, such as for
% the acknowledgment of grants, issue a \IEEEoverridecommandlockouts
% after \documentclass

% for over three affiliations, or if they all won't fit within the width
% of the page, use this alternative format:
% 
%\author{\IEEEauthorblockN{Michael Shell\IEEEauthorrefmark{1},
%Homer Simpson\IEEEauthorrefmark{2},
%James Kirk\IEEEauthorrefmark{3}, 
%Montgomery Scott\IEEEauthorrefmark{3} and
%Eldon Tyrell\IEEEauthorrefmark{4}}
%\IEEEauthorblockA{\IEEEauthorrefmark{1}School of Electrical and Computer Engineering\\
%Georgia Institute of Technology,
%Atlanta, Georgia 30332--0250\\ Email: see http://www.michaelshell.org/contact.html}
%\IEEEauthorblockA{\IEEEauthorrefmark{2}Twentieth Century Fox, Springfield, USA\\
%Email: homer@thesimpsons.com}
%\IEEEauthorblockA{\IEEEauthorrefmark{3}Starfleet Academy, San Francisco, California 96678-2391\\
%Telephone: (800) 555--1212, Fax: (888) 555--1212}
%\IEEEauthorblockA{\IEEEauthorrefmark{4}Tyrell Inc., 123 Replicant Street, Los Angeles, California 90210--4321}}

% use for special paper notices
%\IEEEspecialpapernotice{(Invited Paper)}

% make the title area
\maketitle

\begin{abstract}
\noindent
   Unsupervised image translation aims to learn the transformation from a source domain to a target domain given unpaired training data. Several state-of-the-art works have yielded impressive results in the GANs-based unsupervised image-to-image translation. It fails to capture strong geometric changes between domains, or it produces unsatisfactory results for complex scenes, compared to local texture mapping tasks such as style transfer. Recently, SAGAN \cite{zhang2018self} showed that the self-attention network produces better results than the convolution-based GAN. However, the effectiveness of the self-attention network in unsupervised image-to-image translation tasks have not been verified. In this paper, we propose an unsupervised image-to-image translation with self-attention networks, in which long range dependency helps to not only capture strong geometric change but also generate details using cues from all feature locations. In experiments, we qualitatively and quantitatively show superiority of the proposed method compared to existing state-of-the-art unsupervised image-to-image translation task. The source code and our results are online: \url{https://github.com/itsss/img2img\_sa} and \url{http://itsc.kr/2019/01/24/2019\_img2img\_sa}

\end{abstract}


% For peer review papers, you can put extra information on the cover
% page as needed:
% \ifCLASSOPTIONpeerreview
% \begin{center} \bfseries EDICS Category: 3-BBND \end{center}
% \fi
%
% For peerreview papers, this IEEEtran command inserts a page break and
% creates the second title. It will be ignored for other modes.
\IEEEpeerreviewmaketitle

\section{Introduction}

%\noindent
% Many problems in computer vision and graphics can be posed as an image-to-image translation problem, including inpainting \cite{pathak2016context,iizuka2017globally}, super resolution \cite{dong2016image,kim2016accurate}, colorization  \cite{zhang2016colorful,zhang2017real}, style transfer \cite{gatys2016imag?e,huang2017arbitrary,park2018arbitrary} and so on. This cross-domain image-to-image translation setting has become a major concern of researchers. \\

%\noindent 
In computer vision and graphics there are many image-to-image translation tasks, including inpainting \cite{pathak2016context,iizuka2017globally}, super resolution \cite{dong2016image,kim2016accurate}, colorization  \cite{zhang2016colorful,zhang2017real}, style transfer \cite{gatys2016image,huang2017arbitrary,park2018arbitrary} and so on. This cross-domain image-to-image translation topic has become a major concern of researchers. 

In many cases, given a paired dataset, it is possible to solve the problem with conditional image translation \cite{isola2017image,wang2017high,li2017alice}. However, it is difficult and expensive to obtain the paired samples. In addition, there are cases where supervision is not possible.   

The goal of the unsupervised image translation is to learn the transformation from a source domain to a target domain given unpaired training data. Recent works have yielded impressive results in the GANs-based unsupervised image-to-image translation \cite{yi2017dualgan,zhu2017unpaired,kim2017learning,taigman2016unsupervised,liu2017unsupervised,royer2017xgan,choi2017stargan,huang2018multimodal,anoosheh2017combogan}. It can be largely classified into two types. The first is the style transfer task. This problem is to change low-level information such as color or texture while maintaining high-level information such as content or geometric structure.  Style transfer and conditional GANs-based methods have yielded excellent results in this research area. 

The second is the object transfiguration task. Unlike the style transfer task, this focuses on changing high-level information  while keeping the low-level information. CycleGAN \cite{zhu2017unpaired}, the most representative unsupervised image translation method, failed to change the high-level semantic meaning due to the network structure specialized for style transfer. 

To solve the unsupervised image-to-image translation problem, UNIT \cite{liu2017unsupervised} made a shared-latent space assumption. It assumes a pair of corresponding images in different domains can be mapped to a same latent code in a shared-latent space. MUNIT \cite{huang2018multimodal} proposed a multimodal unsupervised image-to-image translation framework. 

To achieve many-to-many cross domain mapping, it mitigates a fully shared latent space assumption in UNIT by decomposing a shared-latent space across domains and each domain-specific part for the style code.  UNIT and MUNIT experimentally showed impressive animal image translation from a cropped dataset centered on the head. When the training image dataset is spatially unnormalized, it makes the problem more difficult because the absence of correspondences between the shared semantic parts. 

In our experiments, we show that these methods often fail in various image-to-image translation applications with strong geometric change. Recently, SAGAN \cite{zhang2018self} showed that the self-attention module is complementary to convolutions and helps with modeling long range, multi-level dependencies across image regions. Despite the success of the self-attention module in non-conditional GANs, the effectiveness of the self-attention module for unsupervised image-to-image translation has not been validated.  
% Traditional convolutional GANs yields impressive results for the style transfer-like Image-to-image translation tasks (e.g. vangogh2photo, horse2zebra, photo2portrait, summer2winter, etc.) due to few structural constraints. However, it often fails for cross domain translations with geometry or structural changes because these are distinguished not only by local texture but also by global geometry. Recently, SAGAN \cite{zhang2018self} showed that the self-attention module is complementary to convolutions and helps with modeling long range, multi-level dependencies across image regions. Despite the success of the self-attention module in non-conditional GANs, the effectiveness of the self-attention module for unsupervised image-to-image translation has not been validated. 

In this paper, we propose a unpaired image-to-image translation model with self-attention networks which allows long range dependency modeling for image translation task with strong geometry change. In experiments, we show superiority of the proposed method compared to existing state-of-the-art unsupervised image-to-image translation tasks.

The source code and our results are online:
\\
\url{https://github.com/itsss/img2img\_sa} and \url{http://itsc.kr/2019/01/24/2019\_img2img\_sa}.

% Combined with self-attention, the generator can translate images in which fine details at every position are carefully coordinated with fine details in distant portions of the image. Furthermore, the discriminator can also more accurately enforce complicated geometric constraints on the global image structure. 

\section{Self-Attention GANs}

SAGAN \cite{zhang2018self} showed that the self-attention module is complementary to convolutions and it helps with long range modeling, multi-level dependencies across image regions. Attention mechanisms have become a important part of models that must capture global dependencies \cite{cheng2016long, parikh2016decomposable, bahdanau2014neural, xu2015show, yang2016stacked, gregor2015draw}. 

Self attention networks adapt a non-local block \cite{wang2018non} to introduce the self-attention to the GAN networks, can enable both the generator and discriminator to efficiently model relationships between widely separated spatial regions. The non-local mechanisms also have become a important part of image generation \cite{buades2005non, dabov2007image, burger2012image, burger2012image2, glasner2009super, barnes2009patchmatch}. 

\begin{figure}
\begin{center}
\includegraphics[width=\linewidth]{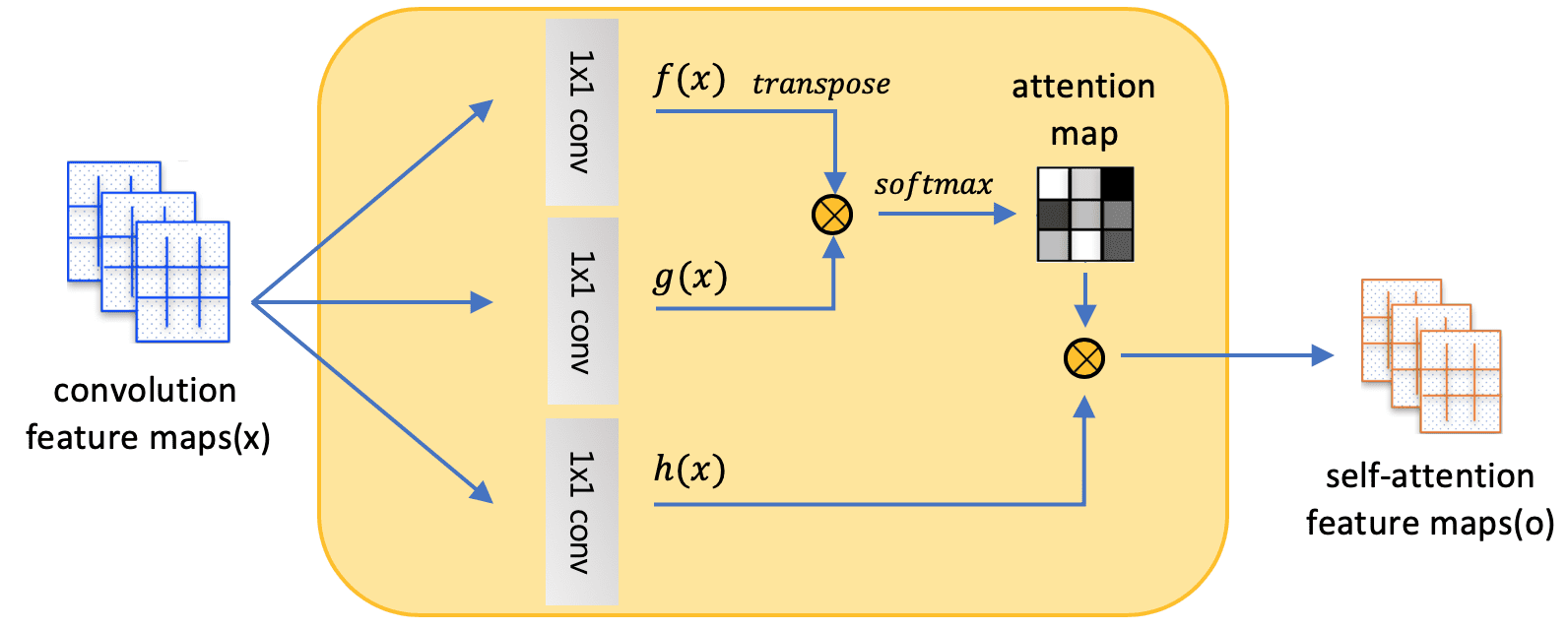}
\end{center}
   \caption{Self Attention Networks. \cite{zhang2018self} $\otimes$ means Matrix multiplication.}
\label{selfattn}
\end{figure}

% \noindent
% In the self attention network model, there are three steps to generate attention layer. \\

% \noindent
% \textbf{Step 1.} Image features from the previous hidden layer $x$ are first transformed into two feature spaces: $f$ and $g$ to calculate the attention, where $f(x) = W_f x$, $g(x) = W_g x$.

% \begin{equation}
%     \beta_{j,i} = \frac{ \exp(s_{ij}) }{ \sum_{i=1}^{N} \exp(s_{ij}) } , where \; s_{ij} = f(x_i)^{T} g(x_j) \nonumber
% \end{equation}

% \noindent
% \textbf{Step 2.} $\beta_{j,i}$ indicated the extent to which the model attends to the $i^{th}$ location when synthesizing the $j^{th}$ region. Then the output of the attention layer is $o = (o_1 , o_2 , ... , o_N)$.

% \begin{equation}
%   o_j = \sum\limits_{i=1}^N \beta_{j,i} h(x_i) \; where \; h(x_i) = W_h x_i \nonumber
% \end{equation}

% \noindent
% \textbf{Step 3.} In the above formulation, $W_g$, $W_f$, $W_h$ are the learned weights matrices, which are implemented as $1 \times 1$ convolutions.

In the self attention module (Figure 1.), image features from the previous hidden layer $x$ are firstly transformed into two feature spaces $f$ and $g$ to calculate the attention. 
\begin{equation}
    \beta_{j,i} = \frac{ \exp(s_{ij}) }{ \sum_{i=1}^{N} \exp(s_{ij}) } , where \; s_{ij} = f(x_i)^{T} g(x_j), \nonumber
\end{equation}
\\
where $f(x) = W_f x$, $g(x) = W_g x$ and $\beta_{j,i}$ indicates the extent to which the model attends to the $i^{th}$ location when synthesizing the $j^{th}$ region. Then the output of the attention layer is $o = (o_1 , o_2 , ... , o_j, ..., o_N)$, where,

\begin{equation}
  o_j = \sum\limits_{i=1}^N \beta_{j,i} h(x_i), \; h(x_i) = W_h x_i \nonumber
\end{equation}
\\
In the above formulation, $W_g$, $W_f$ and $W_h$ are the learned weights parameters, which are implemented as $1 \times 1$ convolutions. 

\section{Methods}
\begin{figure*}
\begin{center}
\includegraphics[width=0.9\linewidth]{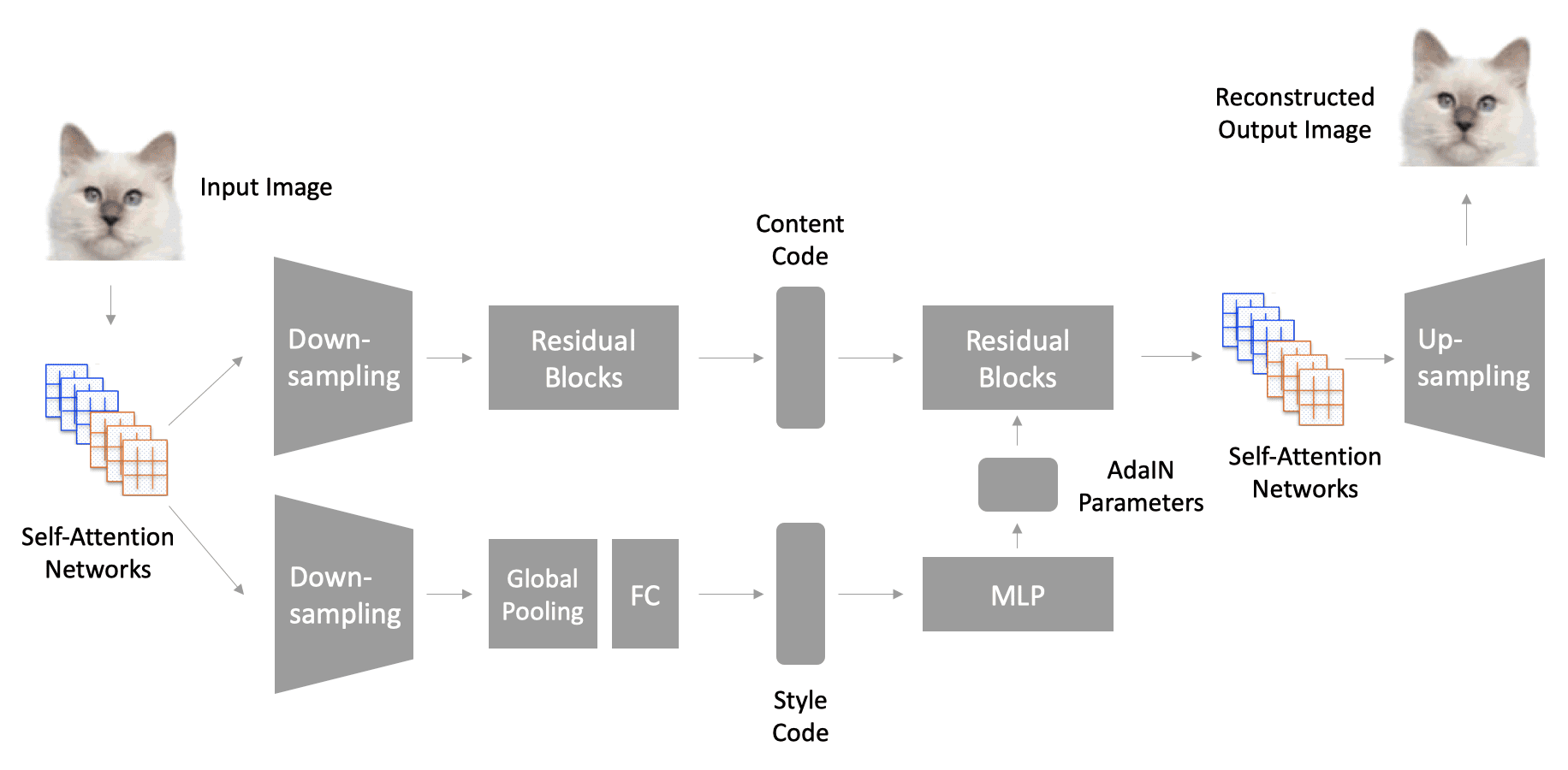}
\end{center}
   \caption{Architecture of our Network Autoencoder Model}
\label{saigan_architecture}
\end{figure*}
\subsection{Unpaired Image-to-Image Translation with Self Attention Networks}

We propose an unsupervised image-to-image translation model with self-attention networks that allows long range dependency modeling for image translation tasks with strong geometry change. Combined with self-attention, the generator can translate images in which fine details at every position are carefully coordinated with fine details in distant portions of the image. Furthermore, the discriminator can also more accurately enforce complicated geometric constraints on the global image structure. 

In this paper, our network architecture is devised by combining several self-attention blocks into the generator and discriminator of the Multimodal Unsupervised Image-to-Image Translation \cite{huang2018multimodal}(MUNIT) model. 

To explore the effect of the proposed self-attention mechanism, we built several SAGAN blocks by adding the self-attention mechanism to different stages of the generator and discriminator. For the generator, the self-attention layers are placed before the downsampling layer in the encoder and before the upsampling layer in the decoder, respectively. For the discriminator, it is added before the downsampling layer. Figure~\ref{saigan_architecture}. shows architecture of our autoencoder model with self-attention networks.

\subsection{Loss Function}

The full objective of our model comprises a bidirectional reconstruction loss function and an adversarial loss function. Same as in \cite{huang2018multimodal}, our model consists of an encoder $E_i$ and a decoder $G_i$ for each domain. The latent code of each autoencoder is divided into a content code $c_i$ and a style code $s_i$, where $(c_i, s_i)$ = $(E^c_i(x_i), E^s_i(x_i))$ = $E_i(x_i)$. Image-to-image translation can be performed by exchanging encoder-decoder pairs. 

\noindent
\textbf{Bidirectional Reconstruction Loss}
Bidirectional reconstruction loss includes image reconstruction loss and latent reconstruction loss. The image reconstruction loss formula is as follows:

\begin{equation}
    \mathcal{L}^{x_1}_{recon} = \mathbb{E}_{x_1\sim p(x_1)} [||G_1 (E^c _1 (x_1), E^s _1 (x_1))-x_1 || _1]. \nonumber
\end{equation}
\\
% Image reconstruction loss given an image sampled from the data distribution, so this loss function be able to reconstruct it after encoding and decoding. The latent reconstruction loss formula is as follows.
We should be able to reconstruct an image sampled from the data distribution after encoding and decoding.

The latent reconstruction loss formula is as follows:

\begin{align}
    \mathcal{L}^{c_1}_{recon} = \mathbb{E}_{c_1\sim p(c_1), s_2\sim q(s_2)} [||E^c _2 (G_2 (c_1, s_2))-c_1 || _1] \nonumber \\
    \mathcal{L}^{s_2}_{recon} = \mathbb{E}_{c_1\sim p(c_1), s_2\sim q(s_2)} [||E^s _2 (G_2 (c_1, s_2))-s_2 || _1] \nonumber
\end{align}
% Latent reconstruction loss given an latent code(content and style) sampled from the latent distribution at translation time, so this loss function should be able to reconstruct it after encoding and decoding. \\
\\
Given a latent code (content an style) from the latent distribution, we should be able to reconstruct it after decoding and encoding. 

\noindent
\textbf{Adversarial Loss}
\noindent
The adversarial loss formula is as follows:

\begin{align}
    \mathcal{L}^{x_2}_{GAN} = \mathbb{E}_{c_1\sim p(c_1), s_2\sim q(s_2)} [\log(1-D_2 ( G_2 ( c_1 , s_2 )))] \nonumber \\ + \mathbb{E}_{x_2\sim p(x_2)} [\log D_2 (x_2)]. \nonumber
\end{align}
\\
% Adversarial loss can employ GANs to the match of the distribution of translated images to the target data distribution. (images generated by network should be indistinguishable from real images in the target domain) \\
\noindent
To match the distribution between the translated and target domain, we employ the adversarial loss.

\noindent
\textbf{Full objective}
The total loss formula is as follows:

\begin{align}
    \min _{ E_1 , E_2 , G_1 , G_2 }{ \max _{ D_1 , D_2 }{ \mathcal{L}(E_1 , E_2 , G_1 , G_2 , D_1 , D_2) }} = \nonumber \\ 
    \mathcal{L}^{x_1}_{GAN} + \mathcal{L}^{x_2}_{GAN} + \lambda_x (\mathcal{L}^{x_1}_{recon} +  \mathcal{L}^{x_2}_{recon} ) + \nonumber \\
    \lambda_c (\mathcal{L}^{c_1}_{recon} + \mathcal{L}^{c_2}_{recon} ) + \lambda_s (\mathcal{L}^{s_1}_{recon} + \mathcal{L}^{s_2}_{recon} ). \nonumber
\end{align} \\

\section{Experimental Results}

In this section, we compared the performance of our model against various unsupervised image-to-image translation models (CycleGAN \cite{zhu2017unpaired}, DRIT \cite{lee2018diverse}, UNIT \cite{liu2017unsupervised}, MUNIT \cite{huang2018multimodal}). In order to evaluate visual quality of translated images, we performed a user study. 

\subsection{Implementation Details}

 We used the MUNIT default setting for experiments. We used the Adam optimizer with $\beta_1 = 0.05$, $\beta_2 = 0.999$. Initial learning rate of 0.0001 and the learning rate is decreased by half every 100,000 iterations. We used a batch size of 1 and set the loss weights to $\lambda_x = 10$, $\lambda_c = 1$, $\lambda_s = 1$. We trained our networks on four TITAN X accelerators. We trained it over 1,000,000 epochs for around 5 days. 

% =============================================================
\begin{figure}
\begin{center}
\includegraphics[width=85.4mm,scale=1]{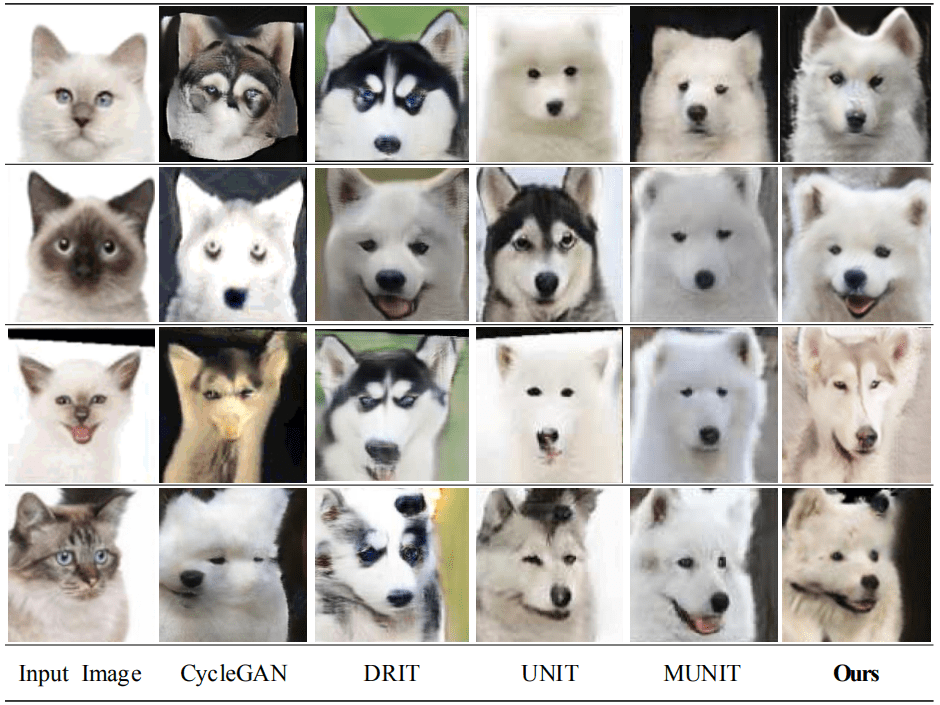}
\end{center}
   \caption{Examples of Unsupervised image translation from cat(cat2dog Dataset, Domain A) to dog(cat2dog Dataset, Domain B) using various network structures. CycleGAN, DRIT, UNIT, MUNIT are all trained to 64 $\times$ 64 resolution using the default settings from the official implementations.}
\label{cat2dog}
\end{figure}

\begin{figure}
\begin{center}
\includegraphics[width=85.4mm,scale=1]{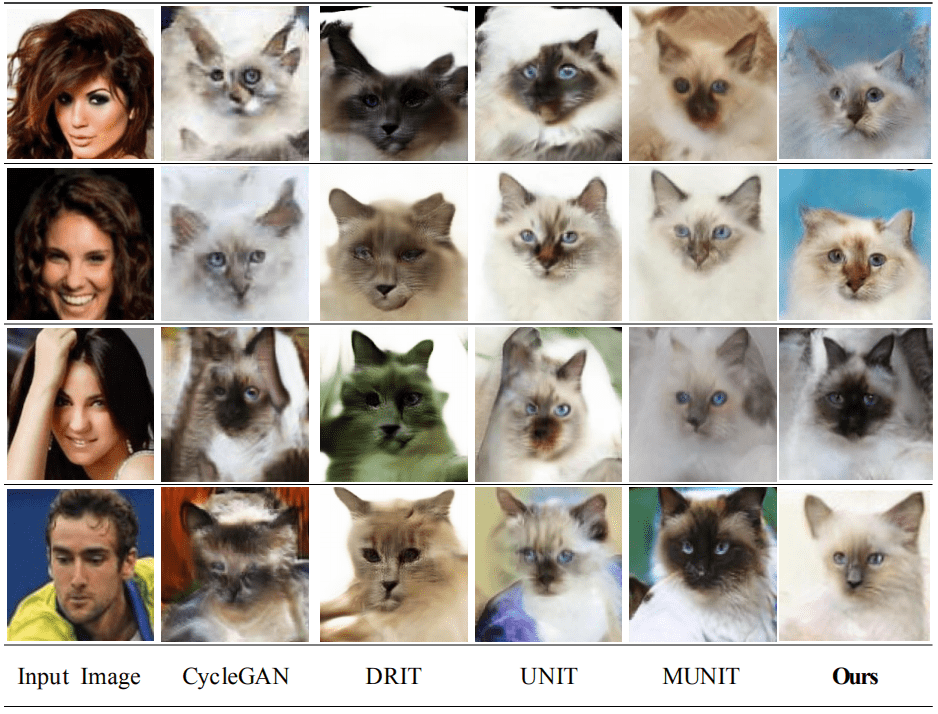}
\end{center}
   \caption{Examples of Unsupervised image translation from human face(CelebA Dataset, Domain A) to cat(cat2dog  Dataset, Domain B) using various network structures. CycleGAN, DRIT, UNIT, MUNIT are all trained to 64 $\times$ 64 resolution using the default settings from the official implementations.}
\label{face2cat}
\end{figure}

% --------------------------------------------------------------

\begin{figure}
\begin{center}
\includegraphics[width=85.4mm,scale=1]{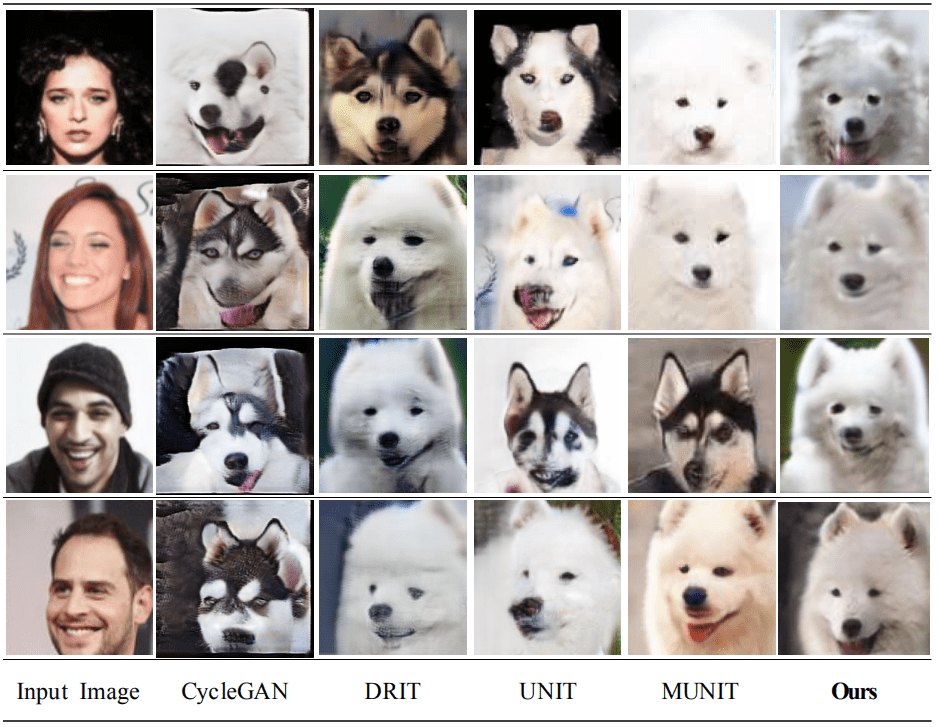}
\end{center}
   \caption{Examples of Unsupervised image translation from human face(CelebA Dataset, Domain A) to dog(cat2dog  Dataset, Domain B) using various network structures. CycleGAN, DRIT, UNIT, MUNIT are all trained to 64 $\times$ 64 resolution using the default settings from the official implementations.}
\label{face2dog}
\end{figure}

\begin{figure}
\begin{center}
\includegraphics[width=85.4mm,scale=1]{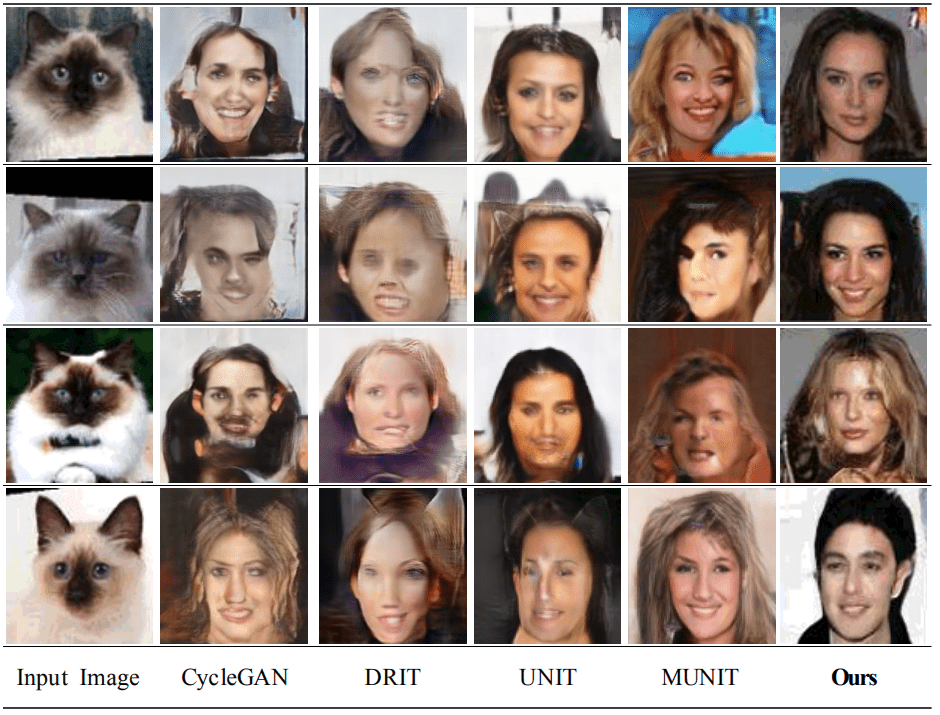}
\end{center}
   \caption{Reverse of Figure~\ref{face2cat}. Examples of Unsupervised image translation from cat(cat2dog Dataset, Domain A) to human(CelebA  Dataset, Domain B) using various network structures. CycleGAN, DRIT, UNIT, MUNIT are all trained to 64 $\times$ 64 resolution using the default settings from the official implementations.}
\label{cat2face}
\end{figure}

% -------------------------------------------------------------
\begin{figure}
\begin{center}
\includegraphics[width=85.4mm,scale=1]{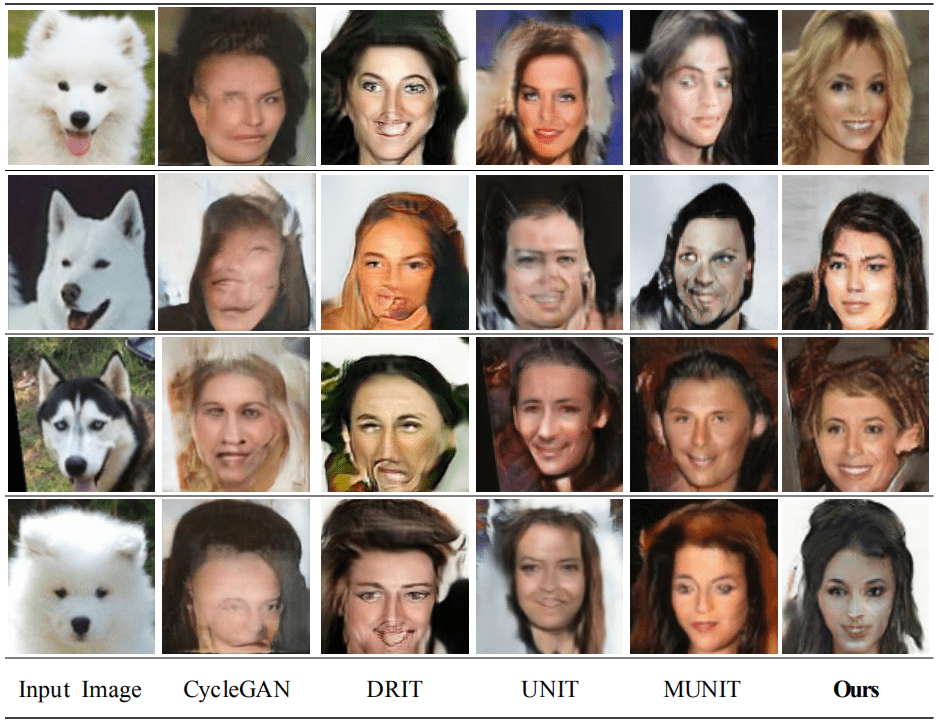}
\end{center}
   \caption{Reverse of Figure~\ref{face2dog}. Examples of Unsupervised image translation from dog(cat2dog Dataset, Domain A) to human(CelebA  Dataset, Domain B) using various network structures. CycleGAN, DRIT, UNIT, MUNIT are all trained to 64 $\times$ 64 resolution using the default settings from the official implementations.}
\label{dog2face}
\end{figure}

\begin{figure}
\begin{center}
\includegraphics[width=85.4mm,scale=1]{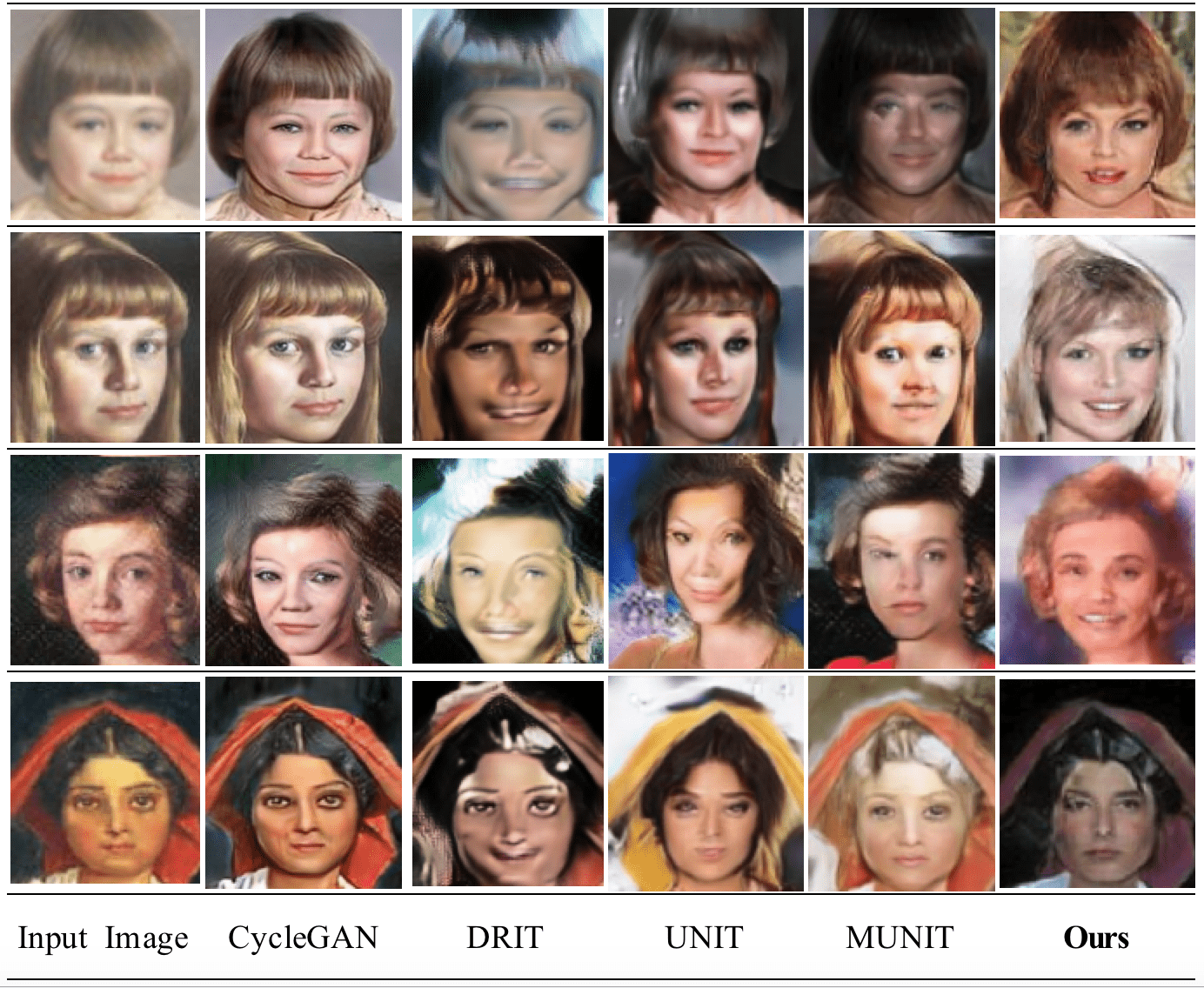}
\end{center}
   \caption{Examples of Unsupervised image translation from portrait(portrait Dataset, Domain A) to human face(portrait Dataset, Domain B) using various network structures. CycleGAN, DRIT, UNIT, MUNIT are all trained to 64 $\times$ 64 resolution using the default settings from the official implementations.}
\label{portrait}
\end{figure}

% -------------------------------------------------------------

\begin{figure}
\begin{center}
\includegraphics[width=85.4mm,scale=1]{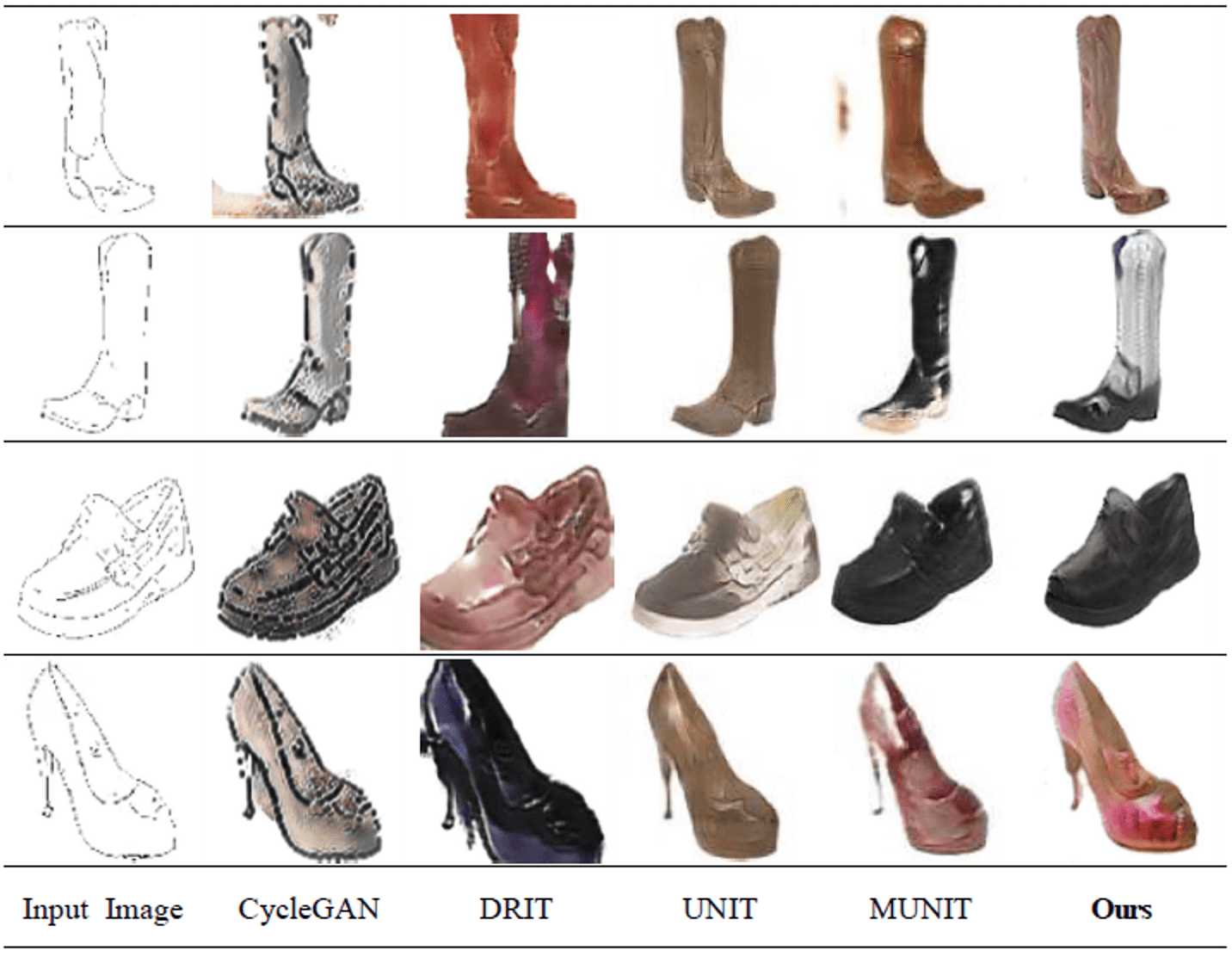}
\end{center}
   \caption{Examples of Unsupervised image translation from edges(edges2shoes Dataset, Domain A) to shoes(edges2shoes Dataset, Domain B) using various network structures. CycleGAN, DRIT, UNIT, MUNIT are all trained to 64 $\times$ 64 resolution using the default settings from the official implementations.}
\label{edges2shoes}
\end{figure}

% =============================================================

\subsection{Datasets}

We used cat2dog, face2dog, face2cat, portrait and edges2shoes for test our network. 

\noindent
\textbf{cat2dog}: This datasets are used in DRIT \cite{lee2018diverse}. This dataset contains cat(871) and dog(1,364). 

\noindent
\textbf{face2dog}: This dataset contains faces (CelebA dataset, 202,599) and dog(cat2dog dataset, 1,364). 

\noindent
\textbf{face2cat}: This dataset contains faces (CelebA dataset, 202,599) and cat(cat2dog dataset, 871).

\noindent
\textbf{portrait}: This datasets are used in DRIT \cite{lee2018diverse}. This dataset contains portrait(1,814) and face photo(6,452). 

\noindent
\textbf{edge2shoes}: This dataset are used in MUNIT \cite{huang2018multimodal}. This dataset contains edges(50,025) and shoes(50,025).

\begin{figure}
\begin{center}
\includegraphics[width=85.4mm,scale=1]{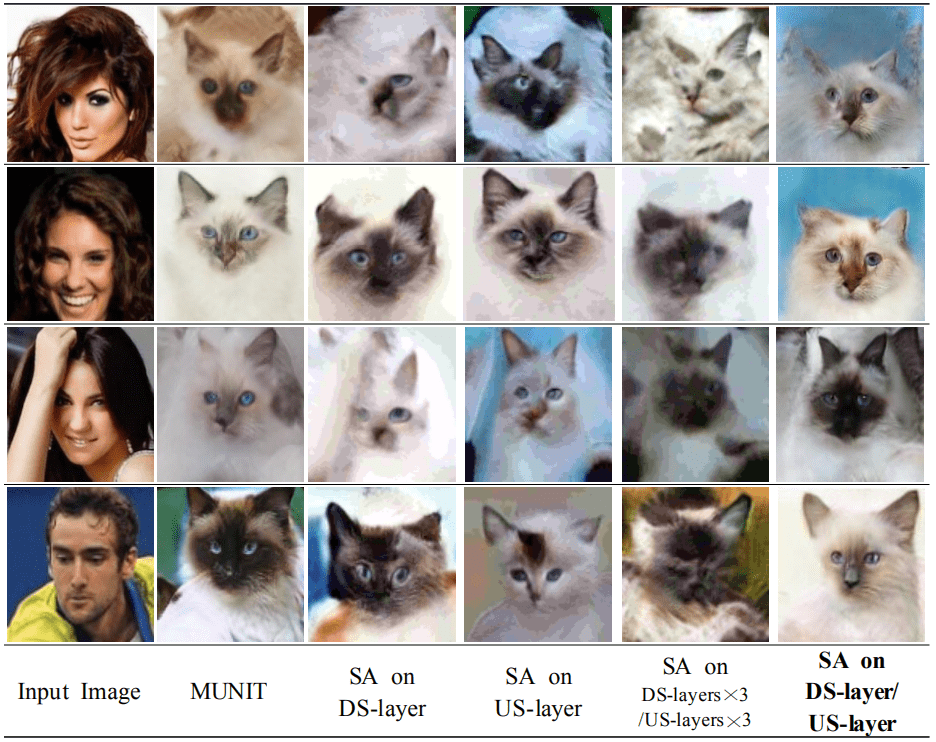}
\end{center}
   \caption{Ablation-study result of four self attention techniques.}
\label{ablation_study}
\end{figure}

\subsection{Comparision with Previous Works}
\noindent
\textbf{cat2dog}
In the process of changing the image of a cat(domain A) to a dog(domain B) image(Figure~\ref{cat2dog}.), CycleGAN is unable to generate a dog image, since it only takes the color from the image. In the case of DRIT, there is a problem that the image is broken, and it is hard to see it as a dog image reflecting the shape and direction of a cat and dog. 
% In the case of UNIT and MUNIT result, the shape and geometry feature was not reflected. 
% In the case of MUNIT, a picture of a totally irrelevant dog was generated.

\noindent
\textbf{face2cat and face2dog}
In the process of translating the human face image(domain A) to cat and dog(domain B), CycleGAN and DRIT could not obtain the desired results. Most translated results were distorted. In the case of UNIT and MUNIT, there was a tendency to leave the shape of human face or be distorted in the translated image (See Figure~\ref{face2cat}. and Figure~\ref{face2dog}.).
% In the case of UNIT and MUNIT, a person's face is kept so badly that it is hard to see a cat coming out.
% In the case of MUNIT, a picture of a totally irrelevant cat was generated.

\noindent
%\textbf{reverse: cat2face, dog2face}
\textbf{cat2face and dog2face}
we also experimented for the translations from cats and dogs to human faces. In this experiments, the proposed method showed much better results than the previous works (See Figure~\ref{cat2face}. and Figure~\ref{dog2face}.).
% We also experimented to convert cats and dogs to humans in order to show that they can be transformed well in the wild image dataset, not just the dataset generated by simply cropping. This result also shows that human face image are properly derived from the gaze, face outline, and appearance of image. all previous works not reflecting shape and contour of cat and dog. The results for reverse can be seen in Figure~\ref{cat2face}. and Figure~\ref{dog2face}. (Domain A: cat and dog, Domain B: human face image)

\noindent
\textbf{portrait}
Even at the stage of changing the portrait (domain A) shown in Figure~\ref{portrait}. to a face(domain B), CycleGAN has not been able to convert portrait photos to face at all. In the case of DRIT, conversion is not performed by generating irrelevant images. In the case of UNIT and MUNIT, there is a problem that the image is distorted although it reflects the shape.

\noindent
\textbf{edges2shoes}
In the process of translating from the edges image(domain A) to a shoes(domain B) image, our model generated more realistic results keeping the pose and style of A domain than the results from other models (See Figure~\ref{edges2shoes}).
% In the process of changing the edges image(domain A) presented in Figure~\ref{edges2shoes}. to a shoes(domain B) image, CycleGAN, DRIT generated image is broken. In the case of UNIT and MUNIT, the model can not create shoes image by keeping only the shape of the photograph. \\

\subsection{Ablation Study}

In this section, the other experiments are conducted to evaluate the effectiveness of the self-attention(SA) networks in our unsupervised image-to-image translation model. In Figure~\ref{ablation_study}. self attention unsupervised image-to-image translation models "SA on downsampling layer (DS-layer)", "SA on upsampling layer (US-layer)" and "SA on DS-layers $\times$ 3 / US-layers $\times$ 3" are compared with a our "SA on DS-layer / US-layer" model.  

In case of "SA on DS-layer", "SA on US-layer" and "SA on DS-layers $\times$ 3 / US-layers $\times$ 3", we could not obtained the well-translated results. However, "SA on DS-layer / US-layer" model generated more realistic images than other methods. Based on this experiments, we applied "SA on DS-layer / US-layer" to our model.

\subsection{User Study}
\begin{figure}[h]
\vskip 0.2in
\begin{center}
\centerline{\includegraphics[width=.6\columnwidth]{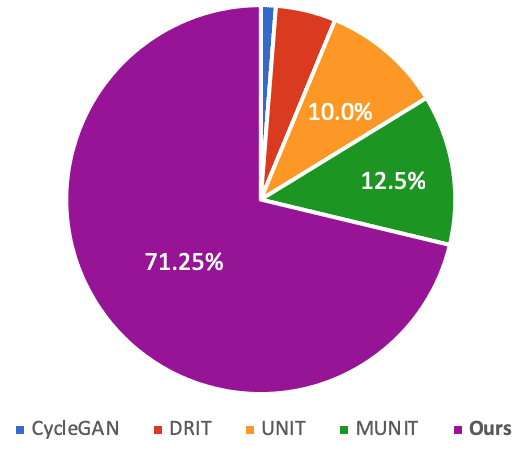}}
\caption{User-study result of five image-to-image translation algorithms.}
\label{user_study}
\end{center}
\vskip -0.2in
\end{figure}

For the qualitative evaluation, we also conducted a user study on 80 participants. The results of this study are summarized as follows. First, 192 images were selected randomly in the questionnaires, and the questionnaires were used to select the best image that reflects the pose of input Image and the appearance of target domain well. Figure~\ref{user_study}. shows that our method yields quantitatively much more superior results than the existing GAN models.

\subsection{Quantitative Evaluation Analysis}
\begin{table}
\begin{center}
\begin{tabular}{|l|c|c|c|c|c|c}
\hline
Dataset & CycleGAN & DRIT & UNIT & MUNIT & Ours \\
\hline\hline
cat2dog & 133.21 & 148.87 & 101.41 & 122.04 & \textbf{96.34} \\
face2cat & 274.61 & 117.05 & 85.78 & 104.09 & \textbf{79.95} \\
face2dog & 279.74 & 108.29 & 82.12 & 133.96 & \textbf{90.70} \\
cat2face & 454.99 & 242.78 & 359.62 & 269.44 & \textbf{208.33} \\
dog2face & 366.33 & 229.21 & 229.06 & 228.06 & \textbf{217.58} \\
portrait & 233.34 & 282.29 & 263.28 & 269.56 & \textbf{256.04} \\
edges2shoes & 269.18 & 273.93 & 250.99 & 274.11 & \textbf{238.57} \\
\hline
\end{tabular}
\end{center}
\caption{Quantitative evaluation on 7 image translation examples. We used Frechet Inception Distance(FID) to measure the performance of various network structures.}
\label{fid}
\end{table}

We used Fréchet Inception Distance (FID) \cite{heusel2017gans} to measure the distance between the data distributions of the source and target domains using the features extracted by the inception networks \cite{szegedy2015going}. The lower FID score indicates that the data distribution of two domains are similar. Table ~\ref{fid}. shows the results of the FID score analysis, and we can see that our model translated more similar images than other image-to-image translation methods.

% To obtain a low FID score, a model needs to generate images that are both high image quality and diversity. We take the generated images and compute the FID score between train data(trainB data) and generated data(AtoB image translation result) If the size of an image is different between the two distributions, we resize images of two distributions same.

\section{Conclusions}

In this paper, we proposed a method about unsupervised image-to-image translation with self-attention networks, in which long range dependency helps to not only capture strong geometric change but also generate details using cues from all feature locations.
In experiments, we showed superiority of the proposed method compared to existing state-of-the-art unsupervised image-to-image translation methods.

\subsection{Acknowledgements}
We would like to thank our sponsors, especially, Sejong Academy of Science and Arts(SASA) in Korea. We also thanks to Artificial Intelligence Research Institute(AIRI) in Korea. This research was funded by the Korea Foundation for the Advancement of Science \& Creativity(KOFAC) Science High School Student R\&E support program.

\bibliographystyle{plain}
\bibliography{ref}

% that's all folks
\end{document}